\documentclass{article}

\usepackage{arxiv}

\usepackage[utf8]{inputenc} 
\usepackage[T1]{fontenc}    
\usepackage{hyperref}       
\usepackage{url}            
\usepackage{booktabs}       
\usepackage{amsfonts}       
\usepackage{nicefrac}       
\usepackage{microtype}      
\usepackage{lipsum}

\title{Regression via Arbitrary Quantile Modeling}
\usepackage{times}
\usepackage{graphicx}
\usepackage{latexsym}

\usepackage{cite}
\usepackage{amsmath,amssymb,amsfonts}
\usepackage{algorithmic}
\usepackage{textcomp}
\usepackage{xcolor}
\usepackage{fancyhdr}
\usepackage{color}
\usepackage{ulem}

\author{Faen Zhang, Xinyu Fan\thanks{Corresponding author}, Hui Xu, Pengcheng Zhou, Yujian He, Junlong Liu\\
\{zhangfaen, fanxinyu, xuhui, zhoupengcheng, heyujian, liujunlong\}@ainnovation.com\\
AInnovation Technology Ltd.} 
  
\begin{document}


\maketitle
\bibliographystyle{my}

\begin{abstract}
  In the regression problem, L1 and L2 are the most commonly used loss functions, which produce mean predictions with different biases. However, the predictions are neither robust nor adequate enough since they only capture a few conditional distributions instead of the whole distribution, especially for small datasets. To address this problem, we proposed arbitrary quantile modeling to regulate the prediction, which achieved better performance compared to traditional loss functions. More specifically, a new distribution regression method, Deep Distribution Regression (DDR), is proposed to estimate arbitrary quantiles of the response variable. Our DDR method consists of two models: a Q model, which predicts the corresponding value for arbitrary quantile, and an F model, which predicts the corresponding quantile for arbitrary value. Furthermore, the duality between Q and F models enables us to design a novel loss function for joint training and perform a dual inference mechanism.  Our experiments demonstrate that our DDR-joint and DDR-disjoint methods outperform previous methods such as AdaBoost, random forest, LightGBM, and neural networks both in terms of mean and quantile prediction.
\end{abstract}

\section{INTRODUCTION}

In recent years, great progress has been made for machine learning thanks to the developments of regression methods, such as AdaBoost, random forest, LightGBM and Neural Network have gained popularity and been widely adopted. 
In each sub-field of machine learning, the regression methods are usually involved. In fact, the loss function plays an important role in the regression method. 
Most regression methods utilize the L1, L2 loss functions to directly obtain mean predictions, such as linear regression and random forest. However, these functions may produce poor results for small datasets due to overfitting. 
To alleviate this problem, many researchers propose to use regularization techniques, like L1-norm~\cite{Tibshirani2011}, 
L2-norm~\cite{Hoerl1970} and dropout~\cite{Srivastava2014}. According to statistical learning theory~\cite{vapnik2013nature}, the large error bound on small datasets could be reduced by data augmentation~\cite{fruhwirth1994data}, multi-task learning~\cite{Caruana1997}, one-shot learning~\cite{LiFei-Fei2006} and transfer learning~\cite{Pan2010}. Here, data augmentation and transfer learning increase training samples size, while multi-task learning utilizes helpful auxiliary supervised information. Despite these considerable advantages, all of these methods require additional datasets or human expertise to support them currently.

Ordinary least-squares regression models the relationship between one or more covariates X and the conditional mean of the response variable Y given X = $x$. Quantile regression~\cite{koenker1978regression,Koenker2001} as introduced by Koenker and Bassett (1978), extends the regression model to the estimation of conditional quantiles of the response variable. The quantiles of the conditional distribution of the response variable are expressed as functions of observed covariates. Since conditional quantile functions completely characterize all that is observable about univariate conditional distributions, they provide a foundation for nonparametric structural models. Quantile regression methods are widely used in many risk-sensitive regression problems, but their performance on small datasets fluctuates like the L1 and L2 loss functions. In economics, food expenditure and household income relationship~\cite{koenker1978regression}, the change of wage structure~\cite{buchinsky1994changes} and many other problem are analyzed with quantile regression. In ecology, quantile regression has been proposed to discover more useful predictive relationships between variables with complex interactions, leading to data with unequal variation of one variable for different ranges, such as growth charts~\cite{wei2006quantile}, prey and predator size relationships~\cite{scharf1998inferring} etc.
The main advantage of quantile regression over ordinary least squares regression is that its flexibility for modeling data with heterogeneous conditional distributions and is more robust to outliers in response measurements.   Moreover, the different measures of central tendency and statistical dispersion are useful for obtaining a more comprehensive analysis of the relationships between variables. For instance, Dunham et al.~\cite{dunham2002influences} analyzed the abundance of lahontan cutthroat trout to the ratio of stream width to depth. With Quantile regression, it indicated a nonlinear, negative relationship with the upper 30\% of cutthroat desities across 13 streams and 7 years. But if just using mean regression exstimates, researchers will mistakenly concluded that there was no relation between trout densities and the ratio of stream width to depth. 

Many traditional learning methods solve quantile regression problems by optimizing quantile loss, such as gradient boosting machine~\cite{friedman2001greedy} and multi-layer perceptron~\cite{gardner1998artificial}. These efforts are intended to predict fixed quantile values, and usually result in linear computation costs compared to the number of fixed quantiles. 
Although we can leverage the structure privilege of multi-layer perceptron to predict multiple quantile values simultaneously, we still cannot predict arbitrary quantile values which is necessary for complicated tasks (e.g. conditional distribution learning). In addition, traditional methods struggled to avoid mathematical illegal phenomenon such as quantile crossing~\cite{Bondell2010,Chernozhukov2010}. This is mainly because these quantile values are either estimated independently.
Since median prediction equals to 50\% quantile prediction, it is natural to utilize a neural network with median regression as backbone    to achieve arbitrary-quantile regression, and then to boost the performances of median and quantile predictions in return. When arbitrary-quantile regression is achieved, quantile values will be estimated universally and continuously, which means we can resolve quantile crossing problems by simply applying gradient regularization. Moreover, contrary to the multi-task learning, the arbitrary-quantile regression does not require extra labels or boost performance through augmenting datasets by enumerating arbitrary quantile inputs. 

In this paper, we propose a Deep Distribution Regression (DDR) mechanism, which consists of a quantile regression model (Q model) and its dual model, cumulative distribution regression model (F model). Both models are deep neural networks, which predict corresponding values of any quantiles and predict quantiles of the corresponding values respectively. The joint training of Q and F provides extra regularization. Extensive experiments demonstrate that DDR outperforms fixed quantile loss and L1 / L2 loss based traditional methods, such as neural networks and ensemble trees. The key contributions of our proposed DDR mechanism include three aspects. First, we design a single neural network for arbitrary quantile prediction, which achieves better quantile loss, L1 loss and even L2 loss compared to fixed-point quantile regression, mean absolute error (MAE) regression and mean squared error (MSE) regression. Second, our DDR method can predict quantile for arbitrary value of corresponding variable, which is useful for anomaly detection or outlier detection. Third, we further utilize joint training and ‘dual inference’ mechanism of the two dual models to obtain better estimation of the whole conditional distribution.

\noindent \paragraph{Contributions} 
The novelties of our method are as follows:
\begin{itemize}

\item We treat quantile as an input variable instead of a fixed constant, which implies DDR uses an integral of loss function as loss function instead of a finite set of loss functions in training process.

\item We introduce mathematical constraint that conditional quantile function should be the inverse function of conditional c.d.f. in DDR, which is not considered in previous studies.

\item We introduce mathematical constraint that conditional quantile function and conditional c.d.f. should both be monotonous in DDR by adding regularization terms on corresponding gradients.

\item We leverage our trained dual models to perform ‘dual inference’ and achieve better performance.
\end{itemize}

\section{METHODOLOGY}
\subsection{Distribution function and quantile function}
Let $X$ be a continuous real-valued random variables with cumulative distribution function (c.d.f.) $F$:
\begin{equation}
F(x)=P(X \leq x),  \qquad x \in \mathbb R
\end{equation}
For each $\tau$ strictly between $0$ and $1$, we define:
\begin{equation}
F^{-1}(\tau)=inf\{x: F(x) \geq \tau\},  \qquad \tau \in (0,1)
\end{equation}
The $F^{-1}(\tau)$ is called the $\tau$ quantile of $X$ or the 100$\tau$ percentile of $X$. The function $F^{-1}$ defined here on the open interval $(0,1)$ is called as quantile function of $X$.

In general cases, the quantile function (q.f.) $Q(\tau)$ is a continuous and strictly increasing function that  takes response variable $X$ as inputs and outputs of the corresponding quantile $q_\tau$:
\begin{equation}
Q(\tau)=inf\{x: F(x) \geq \tau\}=q_\tau,  \qquad \tau \in (0,1)
\end{equation}
hence $q_\tau$ is the $\tau$ quantile of $X$.
Clearly, the q.f. $Q$ and the c.d.f. $F$ are the inverse of each other:
\begin{equation}
Q(F(x))=x 
\end{equation}
\begin{equation}
F(Q(\tau))=\tau
\end{equation}


\subsection{Regression analysis}

In statistical modeling, regression analysis is a set of statistical processes for estimating the relationships among variables. In most cases, the regression analysis estimates the conditional expectation of the dependent variable given a response variable with a mean squared error (MSE) loss function. Meanwhile, there are relatively few studies devoted to the analysis of a certain quantile, and one typical example is estimating the conditional median using the Mean Absolute Error (MAE) loss function. 

In quantile regression, $Q(\tau)$ can be characterized as the unique solution to the problem \cite{peracchi2002estimating}:
\begin{equation}
\underset{x \in \mathbb R}{min}\mathbb E{\ell_\tau}(X-x), 
\end{equation}
where $\ell_\tau$ denotes the asymmetric absolute loss function
\begin{equation}
\ell_\tau(u) = u(\tau - \mathbb I\left\{u<0\right\}), 
\end{equation}
and $\mathbb I\left\{\cdot\right\}$ is the indicator function of the event $A$:
\begin{equation}
\mathbb I\left\{A\right\}=
\begin{cases}
1, & if \quad A \quad is \quad True\\
0, & if \quad A \quad is\quad False
\end{cases} 
\qquad
\forall A \in R
\end{equation}
If $\tau=\frac{1}{2}$,then $\ell_\tau(u)=\frac{|u|}{2} $
and $Q(\tau)$ is the median of $X$.

We seek to find a $\hat{x}$ to minimize
\begin{equation}
\begin{aligned}
 \mathbb E{\ell_\tau}(X-\hat{x}) = (\tau -1) \int_{-\infty}^{\hat{x}}(x-\hat{x})dF(x) 
 +\tau \int_{\hat{x}}^{+\infty}(x-\hat{x})dF(x),
 \end{aligned}
\end{equation}
we have the following form by differentiating with respect to $\hat{x}$, 
\begin{equation}
\begin{aligned}
 0 = (1 - \tau) \int_{-\infty}^{\hat{x}}dF(x) 
 -\tau \int_{\hat{x}}^{+\infty}dF(x) = F(\hat{x}) - \tau
 \end{aligned}
\end{equation}

Considering the problem of estimating the conditional distribution of a scalar random variables $Y$ given a random vector $X = x$ when the available data are a sample from the joint distribution of $(Y,X)$, we can estimate the conditional c.d.f.
\begin{equation}
F(y|x)=P(Y \leq y|X=x),  \qquad y \in \mathbb R
\end{equation}

The other way is to estimate the conditional quantile function (c.q.f.). The $\tau$th c.q.f. has the following form:
\begin{equation}
 Q(\tau|x)= inf\{y: F(y|x) \geq \tau\} ,  \qquad \tau \in (0,1)
\end{equation}
It can be easily verified that 
\begin{equation}
 Q(\tau|x)=F^{-1}(\tau|x),
\end{equation}
\begin{equation}
 F(y|x)=Q^{-1}(y|x),
\end{equation}
The above statements mean that we can achieve estimating conditional expectation when we implement the estimation of arbitrary conditional quantiles, which also shows that an arbitrary quantile modeling is able to cover most regression problems, as evidenced below. Considering that conditional expectation of $Y$ given $X$,
\begin{equation}
E(Y|X=x) = \int_{ - \infty }^{ + \infty } y f_{Y|X}(y|x)dy = \int_{ - \infty }^{ + \infty } ydF(y|x),
\end{equation}
Using the definition of c.q.f. Q and leveraging the integral transformation, we can draw that
\begin{equation}
E(Y|X=x) = \int_{0}^{1}Q(\tau|x)d\tau
\end{equation}
Since it is intractable to calculate the exact integration when the c.q.f. $Q$ has a complex form, we calculate the numerical integration using Trapezoidal rule to estimate the conditional expectation:
\begin{equation}
\begin{aligned}
E(Y|X=x) \approx \frac{\Delta x}{2}(Q(\tau_0|x)+2Q(\tau_1|x)+2Q(\tau_2|x)
+\dots+2Q(\tau_{N-1}|x)+Q(\tau_N|x)), \label{eq1}
\end{aligned}
\end{equation}
where the interval $[0,1]$ is partitioned into $N$ ($N\xrightarrow{}\infty$) equal subintervals, each of width $\Delta x = \frac{1}{N}$, such that $0 = \tau_0  < \tau_1 < \tau_2 < ... < \tau_N = 1$ and $\tau_i = i\Delta x$.

Notice that $Q(\tau_0|x) \sim Q(\tau_N|x)$ can all be calculated with our arbitrary quantile modeling, so conditional expectation can be estimated with equation above, which concludes the proof.
In order to train a quantile model, according to the definition of c.q.f. $Q$,  quantile $\hat{y}$ can be found by minimizing the expected loss of
\begin{equation}
\begin{aligned}
 Q(\tau|x) = arg\underset{\hat{y}}min  \{ (\tau -1) \int_{-\infty}^{\hat{y}}(y-\hat{y}))dF(y|x) 
 +\tau \int_{\hat{y}}^{+\infty}(y-\hat{y})dF(y|x) \}
 \end{aligned}
\end{equation}

When it is actually applied in training process, this formula can be expanded to an empirical form:

\begin{equation}
\ell_{Q(\tau|x)} = \tau |y-Q(\tau|x)|_+ + (1-\tau)|Q(\tau|x)-y|_+ ,
\end{equation}
where $| . |_+$is the ReLU function:
\begin{equation}
|v|_+ = max(v,0), \qquad  \forall v \in R 
\end{equation}

Although some theories and methods have been proposed  to leverage this loss function and estimate quantiles one at a time~\cite{chaudhuri2002nonparametric,chen2016xgboost,He2017,hwang2005simple,Shim2009}, few researchers have considered the panorama of arbitrary conditional quantiles of the response variable. There have been some significant efforts to estimate multiple quantiles in a single neural network with different outputs~\cite{cannon2011quantile,rodrigues2018beyond}, but these methods failed to estimate an infinity set of quantiles (i.e. arbitrary quantile).
To train the conditional c.d.f. regression model, we simply leverage the maximum likelihood estimation method and set the conditional c.d.f. loss to negative likelihood of each batch~\cite{peracchi2002estimating}. We denote $\widetilde{y}$ in the range of $Y$ as the anchor we want to estimate conditional c.d.f. with, and then it is clear that the likelihood could be represented by:  
\begin{equation}
p =   \prod_{y_i \leq \widetilde{y} } F(\widetilde{y}|x) \cdot \prod_{y_i \geq \widetilde{y} }  [1-F(\widetilde{y}|x)], \qquad
i=1,...,J
\end{equation}
Therefore,
\begin{equation}
\begin{aligned}
\log{p} = \sum_{i=1}^J[\mathbb I\{ y_i \leq \tilde{y} \} \cdot \log{ [F(\tilde{y}|x)]+ (1-\mathbb I\{y_i \leq \tilde{y}\})} \cdot \log{[1-F(\tilde{y}|x)]}]\\
= \sum_{i=1}^J [\mathbb I\{ y_i \leq \tilde{y} \} \cdot \log{  \frac{[F(\tilde{y}|x)]}{[1 - F(\tilde{y}|x)]}} + \log{[1-F(\tilde{y}|x)]}
\end{aligned}
\end{equation}
One way to automatically condition $F(\widetilde{y}|x)$ between 0 and 1 is not to directly model $F(\widetilde{y}|x)$ , but rather the log-odds $\eta(\tilde{y}|x) = ln[F(\tilde{y}|x)/(1-F(\tilde{y}|x)]$ \cite{peracchi2002estimating}.
Then $F(\widetilde{y}|x)$ will be estimated by given $\hat{\eta}(\tilde{y}|x)$ using a Sigmoid function $\sigma$:
\begin{equation}
\hat{F}(\tilde{y}|x)=\sigma \big(\hat{\eta}(\tilde{y}|x) \big) = \frac{\exp{ \hat{\eta}(\tilde{y}|x)}}{1+\exp{ \hat{\eta}(\tilde{y}|x)}}
\end{equation}

Therefore, the likelihood could be reformulated as:
\begin{equation}
\log {p}=\mathbb I
\{
y \leq \tilde{y}
\} \cdot 
\eta(\tilde{y}|x)-\ln\big(1+\exp{ \eta(\tilde{y}|x)} \big),
\end{equation}

As a result, the loss function of $F$ model will be:
\begin{equation}
\ell_{F(\tilde{y}|x)}=-\mathbb I
\{y \leq \tilde{y}\} \cdot 
\eta(\tilde{y}|x)+\ln\big(1+\exp{ \eta(\tilde{y}|x)} \big)
\end{equation}
The difference between our proposed DDR mechanism and these mechanisms is how do we deal with the $\tau$ in quantile regression and the $\widetilde{y}$ in the conditional c.d.f. regression, and how we introduce several regularization terms to force $Q(\tau|x)$ and $ F(\widetilde{y}|x) $ to be inverse functions. We will discuss them in detail in \S 3 and describe how our proposed DDR mechanism improves regression performance by simultaneously training $Q$ and $F$ models.

\section{ARBITRARY QUANTILE MODELING}
In this section, we outline the process of building a quantile regression model using the proposed DDR mechanism, which predicts the quantile curves for arbitrary percentile.
For two inverse function $F$ model and $Q$ model mentioned above, we have

\begin{equation}
\begin{cases}
Q(F(y|x)|x) = y, & \forall y \in R,\\ 
F(Q(\tau|x)|x) = \tau, & \forall \tau \in (0,1),
\end{cases} 
\quad
\forall x \in R,
\end{equation}

Similar to neural machine translation, the c.q.f. $Q$ model and the conditional c.d.f. $F$ model can be considered as samples from two model corpora. Unlike neural machine translation, the mapping between these two model corpora is mathematically limited to an inverse mapping, which guaranteed their alignments. This property allows us to train $Q$ model and $F$ model together with some regularization that beyond any aligned data.

In the following, we will firstly introduce how to train a single c.q.f. model and a single conditional c.d.f. model, secondly we will discuss how to add constraints to ensure their inverse mapping. Finally, we will discuss how to use dual inference to potentially boost regression performance.
\subsection{Arbitrary quantile and conditional c.d.f. regression}
As mentioned in \S 2 that quantile regression can be done with some given percentiles, while conditional c.d.f. regression can be done with some given anchors. However, regression with arbitrary percentiles and anchors are not supported by traditional methods, because the latter treat loss functions as `constants' instead of `variables'. Therefore, we will use an integral of corresponding loss function as loss function in our method. 
In DDR mechanism, we model $ Q(\tau,x)$ instead of $Q(\tau|x)$, which implies the Q model is not determined by a specific percentile $\tau$, but takes $\tau$ as input directly.
 In this case, loss function should be an integral of previous loss function:
\begin{equation}
\ell_{Q(\tau,x,y)} = 
min \int _{0}^{1} \tau |y- Q(\tau,x)|_+ + (1-\tau)|Q(\tau,x) - y|_+d\tau.
\end{equation}

Since the integral is intractable, we simply use Monte Carlo method to estimate this loss function in practice. Notice that the different prior distribution assigned to $\tau$ leads to the different ‘attention’ to what we want to settle. In our experiment, we generally sample $\tau$ from uniform distribution U(0,1).
Similarly, we model $F(\widetilde{y},x)$ instead of $F(\widetilde{y}|x)$ using loss function $\ell_{F(\tilde{y}|x)}$. Since the lack of the lower and upper bound of $\widetilde{y}$, we use empirical bounds from our training dataset. We denote x as the original features, y as the labels and D as the dataset, let
\begin{equation}
\begin{cases} 
\widetilde{y}_{min} = min\{ 
y|(x,y) \in D
\} \\
\widetilde{y}_{max} = max\{ 
y|(x,y) \in D
\}
\end{cases}
\end{equation}

In our experiment, we also sample $\widetilde{y}$ from the uniform distribution, but other distribution could be used when more prior knowledge is available.
Since arbitrary $ \tau$ and $\tilde y$ can be seen during training, it is expected that both $Q$ and $F$ model are able to capture quantile curves and conditional c.d.f. curves under arbitrary percentiles and anchors. In practice, however, we may want to focus on specific percentiles and anchors and also ensure their performances, consequently `anchor losses' is introduced to both $Q$ model and $F$ model. We will firstly define two anchor sets for percentiles and conditional c.d.f. anchors that we are interested in:
\begin{equation}
\begin{cases} 
Q_{(anchor)}=
\big\{  
      \tau_1, \dots,\tau_{m}
\big\} \\ 
F_{(anchor)}=
\big\{  
      \widetilde{y}_1, \dots,\widetilde{y}_{m}
\big\}
\end{cases} 
\end{equation}
where $m$ is the number of anchors.

After sampling $\tau$ and $\tilde{y}$ from $U(0,1)$ and $U(\widetilde{y}_{min},\widetilde{y}_{max})$ during training, we will additionally sample $\tau$ and $\tilde{y}$ from $Q^{(anchor)}$ and $F^{(anchor)}$ and then calculate their losses to ensure our $Q$ model and our $F$ model focus more on these anchors.
Moreover, since $Q$ model and $F$ model should satisfy monotonic constraint, we introduce gradient losses as regularization terms as well:
\begin{equation}
\begin{cases}
\ell_g ^{(Q)}(\tau,x) = \sum\limits_{i}{|- \nabla_{\tau}Q(\tau,x_i)|_+
} \\  
\ell_g ^{(F)}(\tilde{y},x) = \sum\limits_{i}{|- \nabla_{\tilde{y}}F(\tilde{y},x_i)|_+
}
\end{cases}
\end{equation}
Notice that $Q$ model and $F$ model should satisfy monotonic constraint not only on inputs from our training set, but also on any inputs sampled from the data distribution. As a result, regularization terms above could be applied to synthetic inputs which are ‘close to’ our dataset, but it never appeared in our dataset (and maybe either in reality). We believe this kind of synthetic regularization will make our model more robust to quantile crossing problem.

\subsection{Recover losses}
For any $\tau \in(0,1)$ and $\tilde{y} \in R $, we introduce two recover losses as follows:
\begin{equation}
\begin{cases}
\ell_r ^{(Q)}(\tau,x) = |\tau - F(Q(\tau,x),x)| \\
\ell_r ^{(F)}(\tilde{y},x) = |\tilde{y}-Q(F(\tilde{y},x),x)|
\end{cases}
\end{equation}
where $\ell_r ^{(Q)}(\cdot)$ means recovering $\tau$ from $Q$ to $F$, and $\ell_r ^{(F)}(\cdot)$ means recovering $\tilde{y}$ from $F$ to $Q$. 
These two losses imply the `inverse mapping' directly, hence forcing $ Q $ and $ F $ to become inverse functions of each other. We simply design absolute error for minimizing the loss.
\subsection{Dual losses}
Apart from recover loss, we also want the mapping between $Q$ and $F$ satisfies their own monotonic constraints mentioned in Section 3.1. 
Specifically, we expect that the $\tau$ recovered from $Q$ to $F$ will still satisfy the monotonic constraint after being input into the $Q$ model again, and the $\tilde{y}$ recovered from $F$ to $Q$ also satisfy the constraint.
Therefore, we introduce two dual losses as follows:
\begin{equation}
\begin{cases}
\ell_d ^{(Q)}(\tau,x) = \sum\limits_{i}{|- \nabla_{\tau}Q(F(Q(\tau,x_i),x_i),x_i)|_+
} \\ 
\ell_d ^{(F)}(\tilde{y},x) =
\sum\limits_{i}{|- \nabla_{\tilde{y}}F(Q(F(\tilde{y},x_i),x_i),x_i)|_+
} \\
\end{cases}
\end{equation}
where $\ell_d ^{(Q)}(\cdot)$ denotes the regularization term on $Q$ by inputting $x$ and recovered $\tau$, and $\ell_d ^{(F)}(\cdot)$ denotes the regularization term on $F$ by inputting  $x$ and recovered $\tilde{y}$. 
These two losses imply that they are homogeneous after the two mappings, hence forcing Q and F to focus more on their own domain.

\subsection{Dual inference}
Since we will train $Q$ and $F$ models simultaneously, when predicting conditional quantile $Q(\tau|x)$, we can either directly calculate it with $Q$ model, or indirectly calculate it by solving an optimization problem with $F$ model:
\begin{equation}
Q(\tau|x) \triangleq \tilde{F}(\tau,x) \triangleq \underset{y}{min}\ell(F(y,x),\tau),
\end{equation}
where loss function $ \ell$ could be a simple L1, L2 loss, or other more specific loss functions when prior knowledge is accessible.

In reality, however, we are not sure whether we obtained a better $Q$ model or we obtained a better $F$ model during the training process, it is natural to output a weighted sum of outputs from both $ Q$ model and $F$ model, which we denoted as dual inference~\cite{xia2017dual}:
\begin{equation}
Q(\tau|x) \triangleq 
\alpha Q(\tau,x)+(1-\alpha)\tilde{F}(\tau,x) \\
\end{equation}
In our experiment, we simply set $\alpha=0.5$, but better performance might be achieved by selecting them according to the performances of $Q$ model and $F$ model on cross validation set.
\subsection{Model structure design and training strategies}
Unlike quantile regression which has different loss functions under different percentiles, median regression holds exclusive loss function, which is known as the L1 loss. Considering that median regression is a special case of quantile regression (where $q_\tau = 0.5$), we build our $Q$ model based on a median regression model. Since every neural network model can be treated as a sequence of transformations from one latent space to another, we inject information of different percentiles by projecting them into those latent spaces and adding them to the original network outputs. 

We will use a two-layer Multi-Layer Perceptron (MLP) model with 256 hidden units and GLU activation function to accomplish the projections, but linear projection or more complex MLP structures could be taken into account when tasks are simpler or more complicated.

In our cases, we will use some structured datasets in experiments, hence network outputs in different latent spaces are simply those activations of different hidden layers in an MLP model. Assume that our MLP model is constructed by a single hidden layer with $l$ neurons, then our median regression model calculates its output by:
\begin{equation}
M(x)=W_2\phi(W_1x+b_1)+b_2, \qquad x\in R^d,
\end{equation}
where  $W_1$ and  $W_2$ are weight matrices with shape $(l, d)$ and $(1,l)$, $b_1$ and $b_2$ are biases vectors with shape $(l,1)$ and $(1, 1)$. Based on this median regression model $M$, our Q model calculates its output by:
\begin{equation}
Q(\tau,x)=W_2\phi(W_1x+b_1+W_\tau \tau+b_\tau)+b_2, \quad \tau \in (0,1),
\end{equation}
where both $W_\tau$ and $b_\tau$ are weight matrix and bias vector with shape of $(l,1)$. Similarly, our conditional c.d.f. regression model calculates its output by:
\begin{equation}
F(\tilde{y},x)=W_2\phi(W_1x+b_1+W_y \tilde{y}+b_y)+b_2, \quad \tilde{y} \in R,
\end{equation}
where $W_y$ and $b_y$ are weight matrix and bias vector both of shape $(l,1)$. These formulas can be extended recursively and thus the information of $\tau$ and $\tilde{y}$ will be directly injected into every latent space in the sequential calculation.

Since our $Q$ model and $F$ model both relied on our median regression model, it is intuitive to primarily train the median regression model to an accepted level before we start to train $Q$ model and $F$ model. Therefore, we introduce some annealing strategies to control the combination of each loss so as to guarantee that we focus more on median regression model in general. Notice that this training strategy could be treated as a hierarchical multi-task (hierarchical ‘infinite-task’ to be exact, since percentiles and anchors are sampled from continuous distributions) learning, which has been proved effective in many domains.

In addition, the proposed (linear) injection of $\tau$'s and $\tilde{y}$'s information means sharing most of the parameters between median regression model, $Q$ model and $F$ model, thus high quality latent features in each latent space are required. Therefore, we separate our MLP model into two parts: the feature part and the regression part. We use GLU activation function and ReLU activation function in feature part and regression part respectively, and update parameters in feature part more frequently than those in regression part.

\section{EXPERIMENTS}
\subsection{Experiments setting}
In the experiments, we will compare the proposed DDR approach with LightGBM~\cite{ke2017lightgbm} and fully-connected neural networks (FCNN) in general quantile regression problems and median regression problems. For the FCNN in general quantile regression problems, we adopt two groups of experiments, namely FCNN and FCNN-joint, to study the effect of shared parameters on quantile regression performance. Specifically, the FCNN method will train k models when we need k quantiles, while the FCNN-joint method will only train 1 model with k outputs to fetch k quantiles together. We will evaluate our algorithm on 6 real datasets and regression benchmarks collected from open sources.

\subsection{Datasets}
As shown in Table \ref{Synthetic}, we firstly evaluated our algorithm on four synthetic 1-dimensional datasets to do sanity checks with clear visualization.
Since the constructed synthetic datasets are rather rough, we also evaluate our methods on the real-world datasets and regression benchmarks collected from open sources,  which are illustrated in Table \ref{Real-world}. The datasets include 6 classes.

\begin{table*}[ht] 
\caption{Synthetic datasets.}\label{Synthetic}
\begin{center}
\begin{tabular}{|l|l|l|l|}
\hline
\textbf{Name}&\multicolumn{1}{|c|}{\textbf{Description}}\\
\hline
linear constant   & 
\begin{tabular}{p{0.4\columnwidth}} linear function with constant noise.\end{tabular} \\
\hline
linear linear   & 
\begin{tabular}{p{0.4\columnwidth}} linear function with noise linear to input.\end{tabular} \\
\hline
quad linear  & 
\begin{tabular}{p{0.4\columnwidth}} quadratic functions with noise linear to input.\end{tabular} \\
\hline
sin constant  & 
\begin{tabular}{p{0.4\columnwidth}} sinusoidal function with constant noise.\end{tabular} \\
\hline
\end{tabular}
\end{center}
\end{table*}

\begin{table*}[ht] 
\caption{Real-world datasets and regression benchmarks collected from open sources.}\label{Real-world}
\begin{center}
\begin{tabular}{|l|l|l|l|}
\hline
\textbf{Name}&\multicolumn{1}{|c|}{\textbf{Sample num}}&\multicolumn{1}{|c|}{\textbf{Feature num}}&\multicolumn{1}{|c|}{\textbf{Description}} \\
\hline
bike  & 17389  & 16  & 
\begin{tabular}{p{0.5\columnwidth}} This dataset contains the hourly and daily count of rental bikes between years 2011 and 2012 in Capital Bikeshare system with the corresponding weather and seasonal information.\end{tabular} 
\\ \hline
year & 515345 & 90 & \begin{tabular}{p{0.5\columnwidth}} Prediction of the release year of a song from audio features. Songs are mostly western, commercial tracks ranging from 1922 to 2011, with a peak in the year 2000s.\end{tabular}                
\\ \hline
fried  & 40768  & 10  & \begin{tabular}{p{0.5\columnwidth}} This is an artificial dataset used in Friedman (1991) and also described in Breiman (1996).\end{tabular}                                                                 \\ \hline
cart  & 40768  & 10  & \begin{tabular}{p{0.5\columnwidth}} This is an artificial dataset similar (but not exactly equal) to the one described in Breiman et al. (1984,p.238).\end{tabular}                                                           \\ \hline
kinematics  & 8192   & 8   & \begin{tabular}{p{0.5\columnwidth}} This dataset is concerned with the forward kinematics of an 8-link robot arm. Among the existing variants of this dataset we have used the variant 8nm, which is known to be highly non-linear and medium noisy.\end{tabular}                                           \\ \hline
\begin{tabular}[c]{@{}l@{}}delta\\  elevators\end{tabular} & 9517   & 6   & \begin{tabular}{p{0.5\columnwidth}} This dataset is obtained from the task of controlling the elevators of a F16 aircraft, although the target variable and attributes are different from the elevators domain. The target variable here is a variation instead of an absolute value, and there is some pre-selection of the attributes.\end{tabular} 
\\ \hline
\end{tabular}
\end{center}
\end{table*}
\subsection{Implementation details}

In order to compare model performance under general quantile regression problems, we decided to compare $10\%, 20\%, \dots, 90\%$ quantile errors together. Here, we trained single model with our DDR mechanism and 9 models based on 9 different quantiles with LightGBM for each task to calculate $q_s$. We denoted $q_s$ as our target metric and it could be calculated by:
\begin{equation}
    q_s \triangleq \sum_{i=1}^9 \ell_Q(0.1*i,x,y).
\end{equation}
We used grid search for LightGBM and developed a set of parameters for DDR which are suitable for various tasks, and we’ll use these parameters throughout this section.

To compare the performance of the model under specific regression problems, we decided to compare median regression performance to the L1 loss and compare the expectation regression performance to the L2 loss.

It should be noticed that the proposed DDR mechanism consists of two main parts: the ‘infinite-task’ learning part and the ‘dual learning’ part. The ‘infinite-task’ learning strategy should boost regression performance by augmenting the data, but it is not clear whether explicitly the impacts of training an additional F model or adding constraints between $Q$ model and $F$ model have positive impacts on the regression performance or not. It is also not clear whether using ‘dual inference’ mechanism will benefit the performance. 

Therefore, we use ‘DDR-q’ to denote DDR mechanism without training $F$ model, ‘DDR-disjoint’ to denote DDR mechanism with F model but without the ‘dual learning’ part and ‘DDR-joint’ to denote DDR mechanism with the ‘dual learning’ part. In the Table \ref{different}, the superscript ‘*’ is used when dual inference with both $Q$ model and $F$ model outperforms direct prediction with $Q$ model on cross validation set (notice that DDR-q is not available to perform dual inference).

\subsection{Experimental results and analysis}

\subsubsection{Results on synthetic dataset}
\begin{figure}[htbp]
\centering
\begin{minipage}[t]{1.0\linewidth}{
\begin{minipage}[t]{0.45\linewidth}
\centering
\includegraphics[width=2.5in]{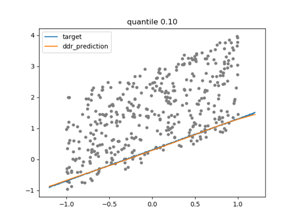}
\end{minipage}
\begin{minipage}[t]{0.45\linewidth}
\centering
\includegraphics[width=2.5in]{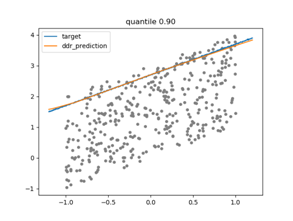}
\end{minipage}}%
\end{minipage}
\centering
\begin{minipage}[t]{1.0\linewidth}{
\begin{minipage}[t]{0.45\linewidth}
\centering
\includegraphics[width=2.5in]{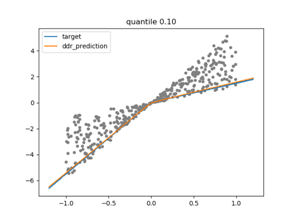}
\end{minipage}%
\centering
\begin{minipage}[t]{0.45\linewidth}
\centering
\includegraphics[width=2.5in]{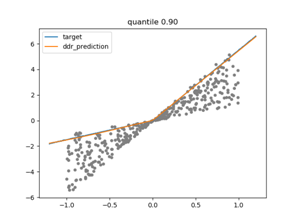}
\end{minipage}}
\end{minipage}
\centering
\begin{minipage}[t]{1.0\linewidth}{
\begin{minipage}[t]{0.45\linewidth}
\centering
\includegraphics[width=2.5in]{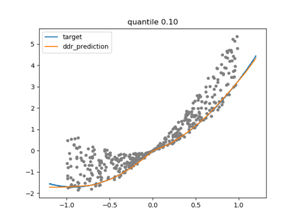}
\end{minipage}%
\centering
\begin{minipage}[t]{0.45\linewidth}
\centering
\includegraphics[width=2.5in]{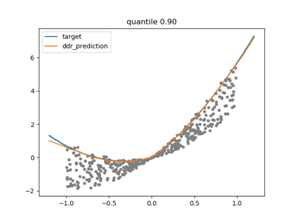}
\end{minipage}}%
\end{minipage}
\centering
\begin{minipage}[t]{1.0\linewidth}{
\begin{minipage}[t]{0.45\linewidth}
\centering
\includegraphics[width=2.5in]{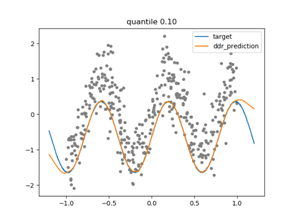}
\end{minipage}%
\centering
\begin{minipage}[t]{0.45\linewidth}
\centering
\includegraphics[width=2.5in]{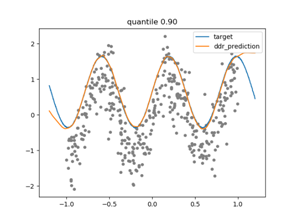}
\end{minipage}}
\end{minipage}
\centering
\caption{Quantile curves of our model's predictions and corresponding ground truths.}\label{fig1} 
\end{figure}

We first experiment on synthetic dataset with 100k samples. Since our synthetic datasets are generated from a certain ground truth function, it is possible for us to get the ground truth quantile curves based on statistic methods (e.g. sampling). In Figure \ref{fig1}, we visualized our model's predictions and corresponding ground truths to prove that our model has enough capacity to learn the inner pattern of the quantile functions.

\subsubsection{Quantile errors on test set and cross validation set}

As shown in Table \ref{different}, dual inference can boost performance in most cases, and DDR-based methods are rather outstanding. It is worth noticing that performance boost on cross validation set could also boost performance on test set in our experiments as shown in Table \ref{testset} and Table  \ref{valset}.

\begin{table*}[htbp]
\caption{Quantile errors of different methods.}\label{different}
\resizebox{\textwidth}{!}{
\begin{tabular}{|l|l|l|l|l|l|l|}
\hline
datasets        & LightGBM & FCNN            & FCNN-joint      & DDR-q           & DDR-disjoint              & DDR-joint                 \\ \hline
bike            & 0.8764   & 0.7254$\pm$0.0847 & 0.6793$\pm$0.1130 & 0.6105$\pm$0.0694 & \textbf{0.5644$\pm$0.0454}  & 0.5999$\pm$0.0315           \\ \hline
year            & 1.9439   & 1.9925$\pm$0.0024 & 1.9171$\pm$0.0082 & 1.9161$\pm$0.0059 & 2.0141$\pm$0.0079*          & \textbf{1.9116$\pm$0.0125*} \\ \hline
fried           & 0.6183   & 0.5675$\pm$0.0049 & 0.5693$\pm$0.0090 & 0.5658$\pm$0.0084 & \textbf{0.5597$\pm$0.0068*} & 0.5603$\pm$0.0021*          \\ \hline
cart            & 0.6569   & 0.6554$\pm$0.0027 & 0.6583$\pm$0.0022 & 0.6551$\pm$0.0014 & \textbf{0.6549$\pm$0.0020*} & 0.6568$\pm$0.0045*          \\ \hline
kinematics      & 1.2911   & 1.1050$\pm$0.0390 & 1.0526$\pm$0.0267 & 0.7472$\pm$0.0256 & 0.7476$\pm$0.0479*          & \textbf{0.7308$\pm$0.0398*} \\ \hline
delta elevators & 1.6546   & 1.6825$\pm$0.0069 & 1.6866$\pm$0.0041 & 1.6475$\pm$0.0043 & 1.6880$\pm$0.0088*          & \textbf{1.6473$\pm$0.0040*} \\ \hline
\end{tabular}}

\end{table*}

More interestingly, the performances of DDR-joint and DDR-disjoint remain consistent between cross validation set and test set as well, which both demonstrate the high generalization ability of our DDR model.

\begin{table*}[ht!]
\caption{Quantile errors of test set.}\label{testset}
\begin{center}
\begin{tabular}{|c|c|c|c|c|}
\hline
& \textbf{joint}      & \textbf{disjoint}  & \textbf{dual-joint}   & \textbf{dual-disjoint}     \\ \hline
bike            & 0.5999$\pm$0.0315          & \textbf{0.5644$\pm$0.0454} & 0.6733$\pm$0.0757          & 0.6685$\pm$0.0743          \\ \hline
year            & 1.9401$\pm$0.0121          & 2.3541$\pm$0.0153          & \textbf{1.9116$\pm$0.0125} & 2.0141$\pm$0.0079          \\ \hline
fried           & 0.5702$\pm$0.0076          & 0.6669$\pm$0.0152          & 0.5603$\pm$0.0021          & \textbf{0.5597$\pm$0.0068} \\ \hline
cart            & 0.6581$\pm$0.0023          & 0.7628$\pm$0.0086          & 0.6568$\pm$0.0045          & \textbf{0.6549$\pm$0.0020} \\ \hline
kinematics      & 0.7593$\pm$0.0406          & 0.9351$\pm$0.0594          & \textbf{0.7308$\pm$0.0398} & 0.7476$\pm$0.0479          \\ \hline
delta elevators & 1.6541$\pm$0.0053          & 1.9399$\pm$0.0298          & \textbf{1.6473$\pm$0.0040} & 1.6880$\pm$0.0088          \\ \hline

\end{tabular}

\end{center}
\end{table*}

\begin{table*}[ht!] 
\caption{Quantile errors of cross validation set.}\label{valset}
\begin{center}
\begin{tabular}{|c|c|c|c|c|}
\hline
                         & \textbf{joint}             & \textbf{disjoint}          & \textbf{dual-joint}        & \textbf{dual-disjoint}     \\ \hline
bike            & 2.7582$\pm$0.1241          & \textbf{2.6494$\pm$0.2045} & 2.8756$\pm$0.0854          & 2.9021$\pm$0.0873          \\ \hline
year          & 1.8897$\pm$0.0042          & 2.3046$\pm$0.0145          & \textbf{1.8630$\pm$0.0062} & 1.9728$\pm$0.0079          \\ \hline
fried           & 0.5718$\pm$0.0037          & 0.6670$\pm$0.0065          & 0.5657$\pm$0.0048          & \textbf{0.5642$\pm$0.0008} \\ \hline
cart            & 0.6368$\pm$0.0034          & 0.7440$\pm$0.0051          & 0.6345$\pm$0.0017          & \textbf{0.6322$\pm$0.0009} \\ \hline
kinematics     & 0.7203$\pm$0.0219          & 0.8850$\pm$0.0624          & \textbf{0.6823$\pm$0.0139} & 0.7028$\pm$0.0586          \\ \hline
delta elevators & 1.5639$\pm$0.0045          & 1.8168$\pm$0.0307          & \textbf{1.5637$\pm$0.0037} & 1.5974$\pm$0.0116          \\ \hline

\end{tabular}
\end{center}
\end{table*}

\subsubsection{Median regression performance}
Apart from the overall performance of quantile regressions, we also compared the median regression performance, which use MAE to measure the performance, between our model and several common methods. The experimental results are illustrated in Table \ref{mae}. In order to compare the performance, we highlight the top 3 performance in the experiments. We find that DDR-joint method still provide better performance for almost all of our experiments.

\begin{table*}[ht!]
	\caption{MAE of the proposed models and commonly used models.} \label{mae}
	\resizebox{\textwidth}{!}{
		\begin{tabular}{|l|l|l|l|l|l|l|}
			\hline
			datasets       & bike            & year                     & fried                    & cart            & kinematics               & delta elevators          \\ \hline
			AdaBoost       & \textbf{0.1352} & 0.5601                   & 0.1948                   & 0.1943          & 0.3832                   & 0.4761                   \\ \hline
			Random Forest  & \textbf{0.1657}          & 0.5714                   & 0.2284                   & \textbf{0.1842}          & 0.4567                   & 0.4694                   \\ \hline
			Neural Network & 0.1943$\pm$0.0289 & 0.5405$\pm$0.0004          & 0.1692$\pm$0.0061          & 0.1856$\pm$0.0005 & 0.3139$\pm$0.0156          & \textbf{0.4613$\pm$0.0021} \\ \hline
			LightGBM       & 0.1755          & 0.5478                   & 0.1742                   & \textbf{0.1842} & 0.3448                   & \textbf{0.4633}                   \\ \hline
			DDR-q          & 0.1723$\pm$0.0078 & \textbf{0.5333$\pm$0.0015}          & \textbf{0.1603$\pm$0.0016}          & 0.1852$\pm$0.0003 & \textbf{0.2064$\pm$0.0068}          & 0.4647$\pm$0.0012          \\ \hline
			DDR-disjoint   & \textbf{0.1550$\pm$0.0156} & \textbf{0.5389$\pm$0.0017}          & \textbf{0.1611$\pm$0.0025}          & 0.1860$\pm$0.0008 & \textbf{0.2103$\pm$0.0142}          & 0.4655$\pm$0.0004          \\ \hline
			DDR-joint      & 0.1667$\pm$0.0180 & \textbf{0.5332$\pm$0.0018} & \textbf{0.1600$\pm$0.0005} & \textbf{0.1851$\pm$0.0009} & \textbf{0.2031$\pm$0.0099} & \textbf{0.4633$\pm$0.0014}          \\ \hline
			
	\end{tabular}}
\end{table*}

\subsubsection{Mean regression performance}
Finally, we also compared the mean regression performance, namely MSE, between our model and several commonly used methods. Note that the mean regression is rather implicit in DDR model, since we can only access it through the equation (\ref{eq1}) mentioned in \S 2.2, where

\begin{equation}
\tau_i=0.01+0.98 \cdot \frac{i}{n+1}.
\end{equation}

We then used this approximation of $E(Y|X)$ to calculate the MSE, and made a comparison to commonly used models which are directly trained with MSE.

From Table \ref{mse}, we can see that the DDR obtains competitive results on MSE metric, even when DDR does not model conditional expectation directly. In order to demonstrate the importance of modeling $Q$ model and $F$ model together, we compare the quantile metric, MAE metric and MSE metric between DDR-disjoint, DDR-joint and DDR-q, where DDR-q only trains $Q$ model and drops out $F$ model.

\begin{table*}[ht!]
	\caption{MSE  of the proposed models and commonly used models.} \label{mse}
	\resizebox{\textwidth}{!}{
		\begin{tabular}{|l|l|l|l|l|l|l|}
			\hline
			datasets       & bike            & year                     & fried                    & cart            & kinematics               & delta elevators          \\ \hline
			AdaBoost       & \textbf{0.0320} & 0.7132                   & 0.0607                   & 0.0601          & 0.2346                   & 0.3910                   \\ \hline
			Random Forest  & 0.0447          & 0.6824                   & 0.0801                   & \textbf{0.0558}          & 0.3101                   & 0.4042                   \\ \hline
			Neural Network & 0.0535$\pm$0.0103 & \textbf{0.6474$\pm$0.0044} & 0.0573$\pm$0.0012          & \textbf{0.0563$\pm$0.0005} & 0.1480$\pm$0.0085          & \textbf{0.3872$\pm$0.0024} \\ \hline
			LightGBM       & 0.0535          & \textbf{0.6760}                   & 0.0463                   & \textbf{0.0558} & 0.1645                   & 0.4060                   \\ \hline
			DDR-q          & \textbf{0.0426$\pm$0.0080} & \textbf{0.6793$\pm$0.0061}          & \textbf{0.0423$\pm$0.0013} & 0.0565$\pm$0.0003 & \textbf{0.0734$\pm$0.0057} & \textbf{0.3886$\pm$0.0019}          \\ \hline
			DDR-disjoint   & 0.0459$\pm$0.0095 & 0.6981$\pm$0.0063          & \textbf{0.0448$\pm$0.0017}          & 0.0573$\pm$0.0007 & \textbf{0.0754$\pm$0.0106}          & 0.4019$\pm$0.0031          \\ \hline
			DDR-joint      & \textbf{0.0436$\pm$0.0090} & 0.6951$\pm$0.0091          & \textbf{0.0414$\pm$0.0013}          & 0.0569$\pm$0.0005 & \textbf{0.0653$\pm$0.0053}          & \textbf{0.3875$\pm$0.002}           \\ \hline
	\end{tabular}}
\end{table*}

\section{CONCLUSIONS}
Comparing to the quantile loss metric and MAE metric, DDR-joint and DDR-disjoint significantly outperforms DDR-q in MSE metric. This is reasonable because we obtain the predictions by numerical integration, which requires general good performances for all possible quantiles at different percentiles ranging from 0 to 1. DDR-joint explicitly models $F(y,x)$ and adds constraints between $Q(\tau,x)$, which may hurt performance at specific percentile (e.g. 0.5 quantile, the median) but boost performance in a more general form.

In summary, we proposes a DDR mechanism to provide a quantile regression, which is a generalization of MAE regression and MSE regression. With this mechanism, we can obtain a model to predict with arbitrary quantile. Moreover, compare with other model, DDR perform better not only in quantile loss, but also L1 loss and L2 loss. Even better estimations are obtained with utilizing dual inference.

In the future, we can continue to explore in the following aspects. Firstly, we have completed experiments on structured dataset and expect to utilize DDR on unstructured dataset, e.g., to replace MSE loss in common regression problems in computer vision such as object detection. Secondly, we can leverage DDR to develop risk prediction models, because our DDR-joint model provides F model which is able to determine whether an observation is abnormal (i.e. whether the F model responses a value close to 0 or 1).


\bibliography{my}

\begin{thebibliography}{10}

\bibitem{Bondell2010}
H.~D. Bondell, B.~J. Reich, and H.~Wang, `Noncrossing quantile regression curve
  estimation', {\em Biometrika}, {\bf 97}(4),  825--838, (dec 2010).

\bibitem{buchinsky1994changes}
Moshe Buchinsky et~al., `Changes in the us wage structure 1963-1987:
  Application of quantile regression', {\em ECONOMETRICA-EVANSTON ILL-}, {\bf
  62},  405--405, (1994).

\bibitem{cannon2011quantile}
Alex~J Cannon, `Quantile regression neural networks: Implementation in r and
  application to precipitation downscaling', {\em Computers \& geosciences},
  {\bf 37}(9),  1277--1284, (2011).

\bibitem{Caruana1997}
Rich Caruana, `{Multitask Learning}', {\em Machine Learning}, {\bf 28}(1),
  41--75, (1997).

\bibitem{chaudhuri2002nonparametric}
Probal Chaudhuri, Wei-Yin Loh, and Others, `Nonparametric estimation of
  conditional quantiles using quantile regression trees', {\em Bernoulli}, {\bf
  8}(5),  561--576, (2002).

\bibitem{chen2016xgboost}
Tianqi Chen and Carlos Guestrin, `{Xgboost: A scalable tree boosting system}',
  in {\em Proceedings of the 22nd acm sigkdd international conference on
  knowledge discovery and data mining}, pp. 785--794. ACM, (2016).

\bibitem{Chernozhukov2010}
Victor Chernozhukov, Iv{\'{a}}n Fern{\'{a}}ndez‐Val, and Alfred Galichon,
  `Quantile and probability curves without crossing', {\em Econometrica}, {\bf
  78}(3),  1093--1125, (may 2010).

\bibitem{dunham2002influences}
Jason~B Dunham, Brian~S Cade, and James~W Terrell, `Influences of spatial and
  temporal variation on fish-habitat relationships defined by regression
  quantiles', {\em Transactions of the American Fisheries Society}, {\bf
  131}(1),  86--98, (2002).

\bibitem{friedman2001greedy}
Jerome~H Friedman, `Greedy function approximation: a gradient boosting
  machine', {\em Annals of statistics},  1189--1232, (2001).

\bibitem{fruhwirth1994data}
Sylvia Fr{\"u}hwirth-Schnatter, `Data augmentation and dynamic linear models',
  {\em Journal of time series analysis}, {\bf 15}(2),  183--202, (1994).

\bibitem{gardner1998artificial}
Matt~W Gardner and SR~Dorling, `Artificial neural networks (the multilayer
  perceptron)—a review of applications in the atmospheric sciences', {\em
  Atmospheric environment}, {\bf 32}(14-15),  2627--2636, (1998).

\bibitem{He2017}
Yaoyao He, Rui Liu, Haiyan Li, Shuo Wang, and Xiaofen Lu, `Short-term power
  load probability density forecasting method using kernel-based support vector
  quantile regression and copula theory', {\em Applied Energy}, {\bf 185},
  254--266, (jan 2017).

\bibitem{Hoerl1970}
Arthur~E. Hoerl and Robert~W. Kennard, `Ridge regression: Biased estimation for
  nonorthogonal problems', {\em Technometrics}, {\bf 12}(1),  55--67, (feb
  1970).

\bibitem{hwang2005simple}
Changha Hwang and Jooyong Shim, `A simple quantile regression via support
  vector machine', in {\em International Conference on Natural Computation},
  pp. 512--520. Springer, (2005).

\bibitem{ke2017lightgbm}
Guolin Ke, Qi~Meng, Thomas Finley, Taifeng Wang, Wei Chen, Weidong Ma, Qiwei
  Ye, and Tie-Yan Liu, `Lightgbm: A highly efficient gradient boosting decision
  tree', in {\em Advances in Neural Information Processing Systems}, pp.
  3146--3154, (2017).

\bibitem{koenker1978regression}
Roger Koenker and Gilbert Bassett~Jr, `Regression quantiles', {\em
  Econometrica: journal of the Econometric Society},  33--50, (1978).

\bibitem{Koenker2001}
Roger Koenker and Kevin~F Hallock, `Quantile regression', {\em Journal of
  Economic Perspectives}, {\bf 15}(4),  143--156, (nov 2001).

\bibitem{LiFei-Fei2006}
{Li Fei-Fei}, R.~Fergus, and P.~Perona, `One-shot learning of object
  categories', {\em IEEE Transactions on Pattern Analysis and Machine
  Intelligence}, {\bf 28}(4),  594--611, (apr 2006).

\bibitem{Pan2010}
Sinno~Jialin Pan and Qiang Yang, `{A Survey on Transfer Learning}', {\em IEEE
  Transactions on Knowledge and Data Engineering}, {\bf 22}(10),  1345--1359,
  (oct 2010).

\bibitem{peracchi2002estimating}
Franco Peracchi, `On estimating conditional quantiles and distribution
  functions', {\em Computational statistics \& data analysis}, {\bf 38}(4),
  433--447, (2002).

\bibitem{rodrigues2018beyond}
Filipe Rodrigues and Francisco~C Pereira, `Beyond expectation: Deep joint mean
  and quantile regression for spatio-temporal problems', {\em arXiv preprint
  arXiv:1808.08798}, (2018).

\bibitem{scharf1998inferring}
Frederick~S Scharf, Francis Juanes, and Michael Sutherland, `Inferring
  ecological relationships from the edges of scatter diagrams: comparison of
  regression techniques', {\em Ecology}, {\bf 79}(2),  448--460, (1998).

\bibitem{Shim2009}
Jooyong Shim and Changha Hwang, `Support vector censored quantile regression
  under random censoring', {\em Computational Statistics {\&} Data Analysis},
  {\bf 53}(4),  912--919, (feb 2009).

\bibitem{Srivastava2014}
Nitish Srivastava, Geoffrey Hinton, Alex Krizhevsky, Ilya Sutskever, and Ruslan
  Salakhutdinov, `Dropout: A simple way to prevent neural networks from
  overfitting', {\em Journal of Machine Learning Research}, {\bf 15},
  1929--1958, (2014).

\bibitem{Tibshirani2011}
Robert Tibshirani, `Regression shrinkage and selection via the lasso: a
  retrospective', {\em Journal of the Royal Statistical Society: Series B
  (Statistical Methodology)}, {\bf 73}(3),  273--282, (jun 2011).

\bibitem{vapnik2013nature}
Vladimir Vapnik, {\em The nature of statistical learning theory}, Springer
  science \& business media, 2013.

\bibitem{wei2006quantile}
Ying Wei, Anneli Pere, Roger Koenker, and Xuming He, `Quantile regression
  methods for reference growth charts', {\em Statistics in medicine}, {\bf
  25}(8),  1369--1382, (2006).

\bibitem{xia2017dual}
Yingce Xia, Jiang Bian, Tao Qin, Nenghai Yu, and Tie-Yan Liu, `Dual inference
  for machine learning.', in {\em IJCAI}, pp. 3112--3118, (2017).

\end{thebibliography}
\end{document}